\documentclass[letterpaper, 10 pt, conference]{ieeeconf}  

\IEEEoverridecommandlockouts                              

\usepackage{graphics} 
\usepackage{epsfig} 
\usepackage{mathptmx} 
\usepackage{times} 
\usepackage{amsmath} 
\usepackage{amssymb}  
\usepackage{booktabs}
\usepackage[font=small,belowskip=-1.5pt]{caption}

\title{SCP: Soft Conditional Prompt Learning for Aerial Video Action Recognition}

\author{Xijun Wang*$^{1}$, Ruiqi Xian*$^{2}$, Tianrui Guan$^{1}$, Fuxiao Liu$^{1}$ and Dinesh Manocha$^{1}$
\thanks{*These authors contributed equally}
\thanks{$^{1}$Authors are with Dept. of Computer Science, University of Maryland, College Park, MD, USA.
        {\tt\small xijun@umd.edu}}%
\thanks{$^{2}$Author is with the Dept. of Electrical and Computer Engineering, University of Maryland, College Park, MD, USA
        {\tt\small rxian@umd.edu}}%
}

\begin{document}

\maketitle
\thispagestyle{empty}
\pagestyle{empty}
\begin{abstract}
  
  We present a new learning approach, Soft Conditional Prompt Learning (SCP), which leverages the strengths of prompt learning for aerial video action recognition.
  Our approach is designed to predict the action of each agent by helping the models focus on the descriptions or instructions associated with actions in the input videos 
  for aerial/robot visual perception. Our formulation supports various prompts, including learnable prompts, auxiliary visual information, and large vision models to improve the recognition performance. We present a soft conditional prompt method that learns to dynamically generate prompts from a pool of prompt experts under different video inputs. By sharing the same objective with the task, our proposed SCP can optimize prompts that guide the model's predictions while explicitly learning input-invariant (prompt experts pool) and input-specific (data-dependent) prompt knowledge. In practice, we observe a $3.17-10.2\%$ accuracy improvement on the aerial video datasets (Okutama~\cite{barekatain2017okutama}, NECDrone~\cite{choi2020unsupervised}), which consist of scenes with single-agent and multi-agent actions. We further evaluate our approach on ground camera videos to verify the effectiveness and generalization and achieve a $1.0-3.6\%$ improvement on SSV2~\cite{goyal2017something}. We integrate our method into the ROS2 as well.

\end{abstract}

\section{Introduction}

\begin{figure}[t]
    \centering
    \includegraphics[width=\columnwidth]{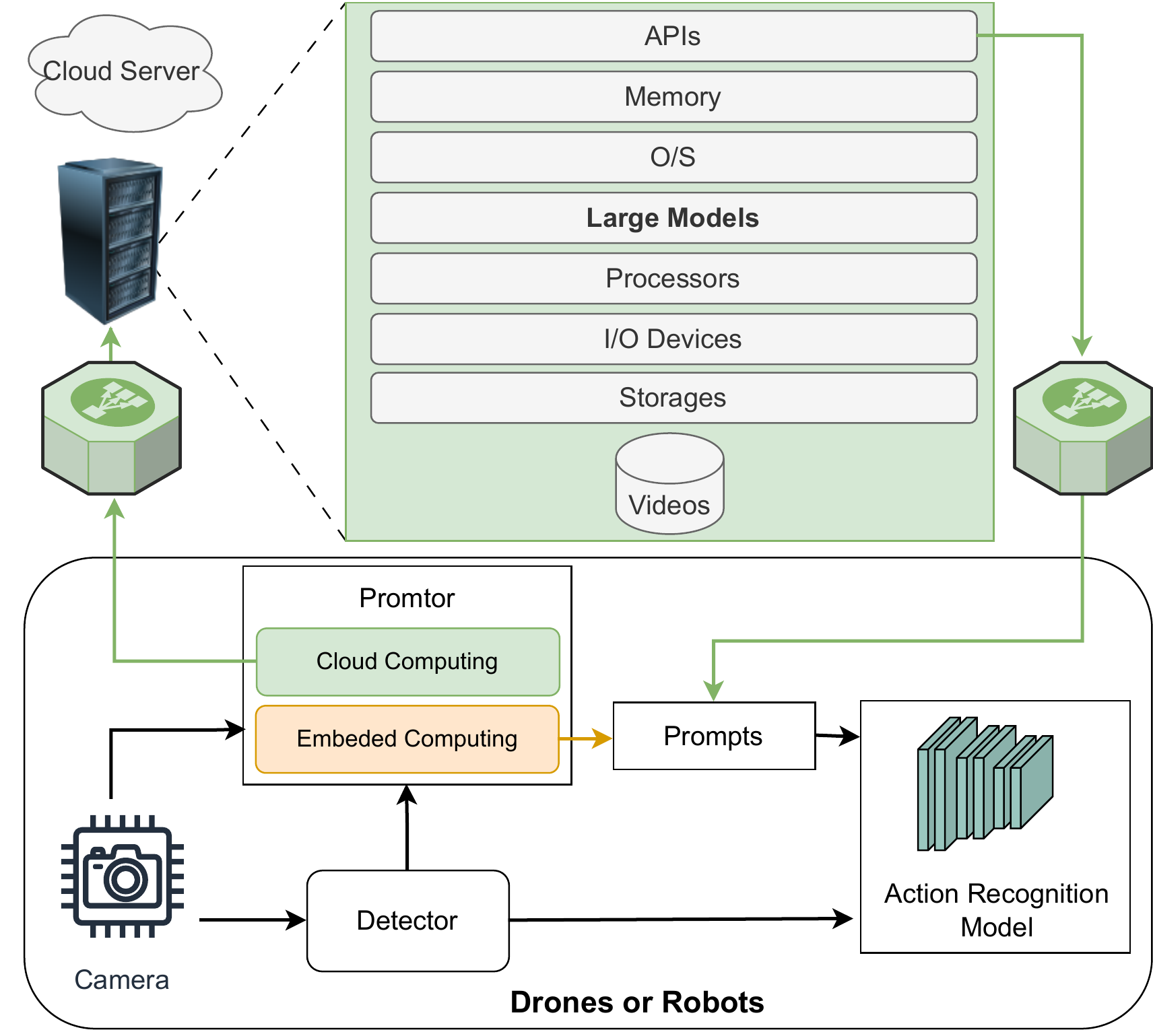}
    \caption{{\bf Overall Architecture:} Our action recognition method is designed to run one edge devices (on mobile robots) and cloud servers. This includes lightweight prompts (embedded), which can be easily embedded in any action recognition model without much extra computational cost. 
    For large vision models, we perform these computations on cloud server and use low-latency communication with the robots. 
    }
    \label{fig:sys} 
    \vspace{-6mm}
\end{figure}

In the realm of unmanned aerial vehicles (UAVs), the ability to accurately recognize human actions from video footage is paramount for safe and effective operation. This entails extracting meaningful insights into the activities and movements of people and objects within the environment, leveraging video sequences captured by the onboard camera. This capability serves as a cornerstone technology for UAV applications, including human-UAV interaction, search and rescue, and comprehensive aerial surveillance.

Similar to ground robots, recent advancements in deep learning techniques have yielded significant strides in human action recognition for UAV videos. However, a crucial challenge persists: the majority of existing approaches rely heavily on extensive, meticulously labeled training datasets and adhere to a purely supervised learning paradigm that primarily focuses on optimizing the network architecture design. When directly applied to aerial video datasets, these methods often experience significant degradation in performance due to inherent challenges unique to this domain including small target sizes, disparate viewing angles, and camera motion dynamics.



\begin{figure*}[t]
    \centering
    \includegraphics[width=\textwidth]{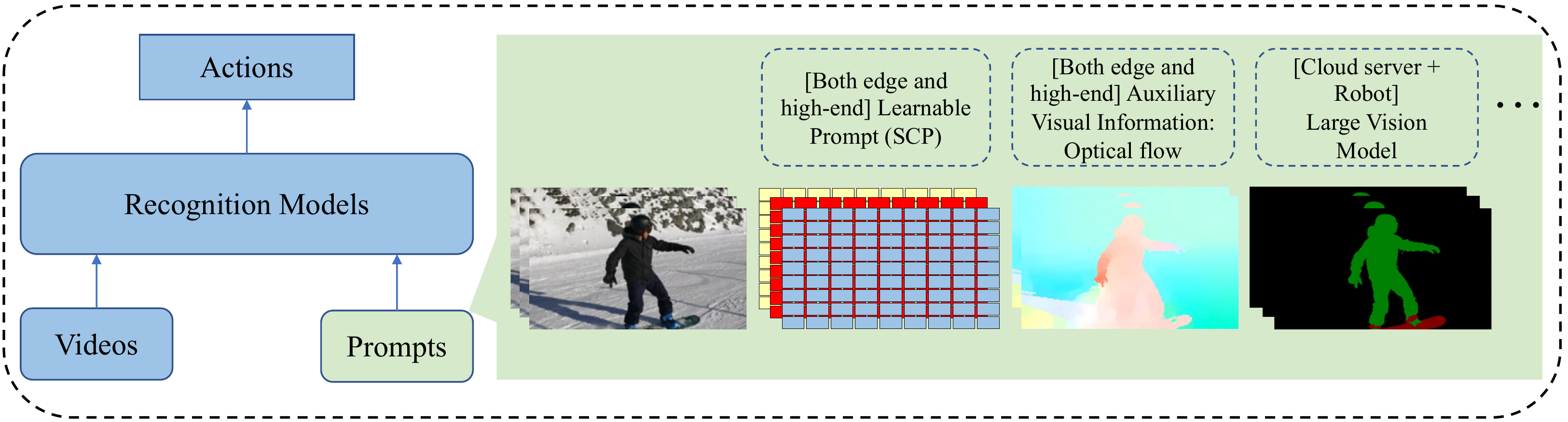}
    \caption{{\bf Task Overview:} We use prompt learning for action recognition. Our method leverages the strengths of prompt learning to guide the learning process by helping models better focus on the descriptions or instructions associated with actions in the input videos. We explore various prompts, including optical flow, large vision models, and proposed SCP to improve recognition performance. The recognition models can be CNNs or Transformers.}
    \label{fig:task} 
    \vspace{-6mm}
\end{figure*}

To address these limitations, our work explores the application of prompt learning for UAV video action recognition. Prompt-based learning techniques, recently demonstrating success in natural language processing tasks~\cite{liu2023pre}, circumvent the requirement for extensive labeled data by leveraging pre-trained language models. In the context of UAV action recognition, prompt learning offers a promising avenue for designing more robust recognition models. By incorporating high-level texture descriptions or instructions associated with actions, prompts can effectively guide the model's learning process. This targeted guidance allows the model to focus on discriminative spatiotemporal patterns in the aerial video data, especially when dealing with challenging visual features like small targets or unusual camera angles. Furthermore, the ease of obtaining or embedding prompt information within existing robotic systems facilitates the practical implementation of this approach.

\noindent {\bf Main Results:} 
In this paper, we propose a novel prompt-learning approach to address the challenges of UAV video action recognition. Our approach integrates prompts to enhance the model's ability to process video data effectively. These prompts can be either learnable or pre-defined templates specifically designed for action recognition tasks. By incorporating prompts, our method facilitates the model's focus on critical regions of interest within video frames. This targeted focus enables the learning of complex visual concepts, such as recognizing interactions between multiple agents in aerial footage.

In our prompt learning paradigm, we explore and discuss different types of prompts, including learnable prompts, auxiliary visual information (optical flow, detection, etc.), and large vision models. For learnable prompts, our SCP dynamically generates prompts from a pool of prompt experts under different inputs. Our goal is to optimize prompts that guide the model's predictions while explicitly learning input-invariant (prompt experts) and input-specific (data-dependent) prompt knowledge. For auxiliary visual information, we can easily obtain them from the robot's built-in system. 
Our SCP can be easily embedded in any model without much extra computational cost, especially suitable for edge and mobile devices.
We validate the generalization by performing evaluations on datasets comprised of aerial videos and ground camera videos on scenarios involving single-agent and multi-agent actions. We demonstrate that our technique can improve performance and enhance the generalization capabilities of video action recognition models in different scenarios. Our main contributions include:

\begin{enumerate}
    \item We present a general learning approach to use prompt learning and auto-regressive techniques for aerial video action recognition. 
    \item We propose a new soft conditional learnable prompt method that can guide the model's predictions while explicitly learning input-invariant (prompt experts) and input-specific (data-dependent) prompt knowledge.
    \item To the best of our knowledge, ours is the first approach to explore the possibility of using large vision models as the prompt to instruct the models on aerial video action recognition tasks.
    \item Through empirical evaluations, we demonstrate the potential and effectiveness of prompt learning techniques for aerial video action recognition tasks. Specifically, we observe a 3.17-10.2\% accuracy improvement on the aerial video datasets. Moreover, we observe a 1.0-3.6\% accuracy improvement on the ground camera video dataset Something Something V2. 
\end{enumerate}

\section{Related Works}
\subsection{Action Recognition}
Human action recognition, i.e., recognizing and understanding human actions, is crucial for a number of real-world applications. Recently,  many deep learning architectures have been proposed to improve the performance. At a broad level, they can be classified into three categories: Two-stream 2D Convolutional Neural Network~\cite{simonyan2014two,karpathy2014large,wang2015action,sanchez2013image,cheron2015p,girdhar2017actionvlad}, 3D CNN-based methods~\cite{tran2015learning,ji20123d,zhang2020few,li2020directional,carreira2017quo}, Transformer-based approaches~\cite{arnab2021vivit, bertasius2021space,wang2021oadtr}.

\begin{figure*}[!h]
    \centering
    \includegraphics[width=\textwidth]{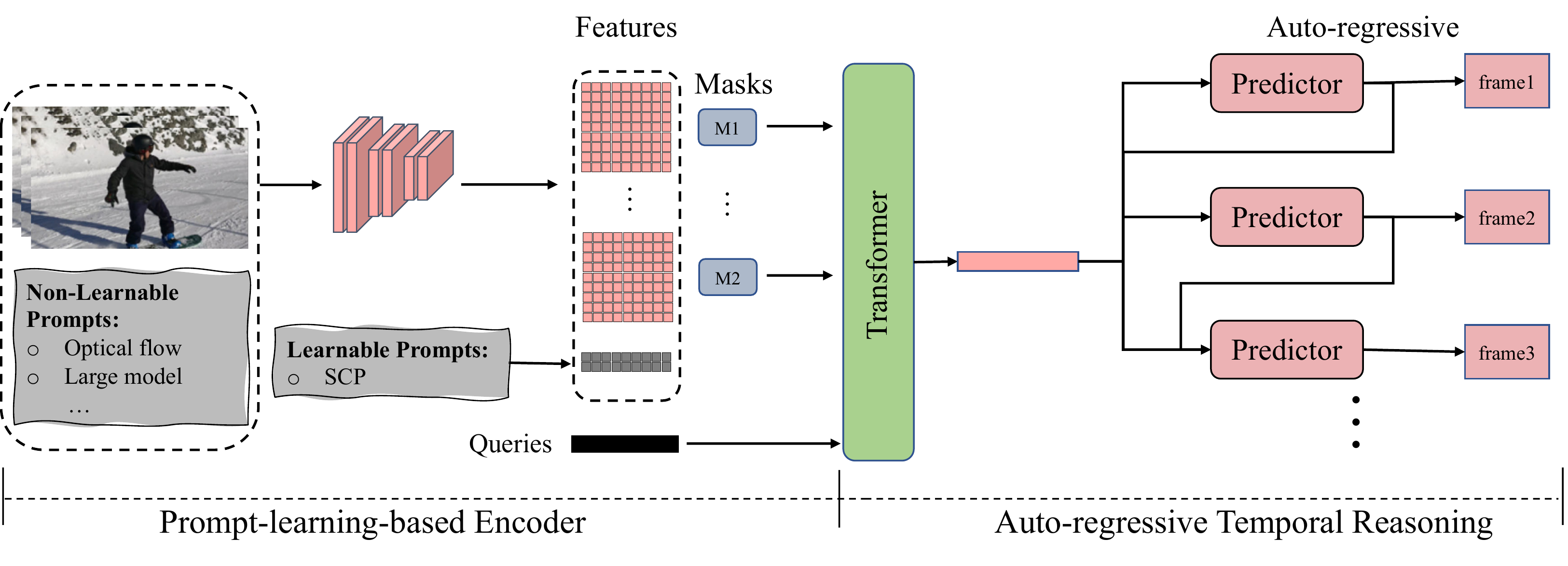}
    \caption{{\bf Overview of the action recognition framework:} We use transformer-based action recognition methods as an example. We designed a prompt-learning-based encoder to help better extract the feature and use our auto-regressive temporal reasoning algorithm for recognition models for enhanced inference ability. }
    \label{fig:overview_method} 
    \vspace{-6mm}
\end{figure*}

Although these methods have had good success on the ground data and YouTube videos, they cannot achieve a similar level of accuracy on videos captured using Unmanned Aerial Vehicles (UAVs)~\cite{wang2023aztr,xian2023pmi}. Compared to ground or YouTube videos, UAV videos have unique characteristics like small resolution, scale and size variations, and moving cameras. 
\cite{wang2023aztr} proposed auto zoom algorithms with an attention mechanism for inference on both edge devices and desktop GPUs. 
\cite{xian2023mitfas} proposed a mutual information-based feature alignment and sampling method to extract spatial-temporal features corresponding to human actors for better recognition accuracy. \cite{Kothandaraman2022FARFA} introduced Fourier transformation into attention modules to aggregate the motion salience. \cite{xian2023pmi} proposed a novel frame sampler for aerial action recognition by measuring the similarity between frame patches. Our SCP can help the above methods better focus on the target agents.

\subsection{Prompt Learning}
The concept of prompt learning, initially introduced by \cite{petroni2019language}, has garnered significant attention in the field of Natural Language Processing (NLP)\cite{brown2020language,jiang2020can,Li2021PrefixTuningOC,liu2023pre,tian2020rethinking,li2023towards}. Prompt learning revolves around the fundamental idea of treating pre-trained language models like BERT or GPT as knowledge repositories, enabling their utilization in downstream tasks. Early studies, exemplified by\cite{petroni2019language,Poerner2019EBERTEE}, concentrated on crafting prompts manually to enhance language model performance. Subsequently, researchers like \cite{Shin2020ElicitingKF,jiang2020can} aimed to automate this process using cost-effective, data-driven approaches. More recently, some works\cite{han2022ptr,lester2021power,zhong2021factual} have ventured into learning continuous prompts as an alternative to seeking discrete prompts.


In \cite{jiang2022vima}, the versatility of expressing a wide range of robot manipulation tasks through multimodal prompts is demonstrated using VIMA, a transformer-based generalist robot agent that processes prompts and generates motor actions autoregressively. \cite{singh2023progprompt} introduces a programmatic LLM prompt structure to facilitate plan generation adaptable to various settings, robot functionalities, and tasks. Additionally, \cite{vemprala2023chatgpt} proposes a strategy combining prompt engineering principles and a high-level function library to enhance ChatGPT's adaptability to diverse robotics tasks, simulation environments, and hardware setups. In fashion, ~\cite{wang2024icar} use scenes as prompts to help style-matched recommendations.  In foundation model design, ~\cite{wang2023scsc} explores different spatial information as prompts. ~\cite{wang2023vlap,Wang2024} explore how to find the keyframes efficiently as prompts for LLMs. Recently, more and more researchers started exploring prompt learning techniques in vision tasks~\cite{rao2022denseclip,ju2022prompting,Zhou2021LearningTP,liu2023aligning,liu2023mmc}. 

While previous research has predominantly concentrated on prompt learning for ground robot tasks, the application of prompt learning to UAV tasks has received limited attention. This paper introduces a comprehensive learning framework aimed at assessing the efficacy of prompt learning in the context of UAV video comprehension, particularly in the realm of action recognition in both ground/YouTube and aerial videos. The objective is to bridge this gap and broaden the applicability of prompt learning to video understanding tasks within this domain.

\section{Our Approach}

We denote the input as $X_i=\{x_1, x_2, ..., x_m\}, i\in [1, N]$, where $x_j$ is the $j_{th}$ frame in the $i_{th}$ video, $m$ is the total frame number, and $N$ is the total number of videos. The overall approach predicts the action categories by using model $f(X_i)$, which can be CNNs or Transformers. As shown in Figure~\ref{fig:overview_method}, taking transformer-based methods as an example, we follow the same scheme to extract the features, followed by using the reasoning process to predict the action labels. We also present a prompt-learning-based encoder to help better extract the feature and then propose an auto-regressive temporal reasoning algorithm for recognition models for enhanced inference ability. Specifically, in an action model:
 \begin{equation}
 f=f_a\circ f_e([X,P]), 
\end{equation}
 where $f_e$ is the \emph{prompt-learning-based input encoder}, $P$ is the prompt, and $f_a$ is the \emph{auto-regressive-based temporal reasoning} model, which is used for the temporal dimension.

\subsection{Prompt Learning-based Input Encoder}
For the first part of the input encoder, inspired by these prompt-based techniques in NLP, we present a new general prompt learning-based input encoder for action recognition. Our formulation leverages the strengths of prompt learning to guide the optimization by providing high-level descriptions or instructions associated with actions in the inputs. We use this to alleviate the burden of models' optimization by helping models better focus on the active region. 
 
Prompts can enhance the model's ability to process customized inputs by utilizing prompt tokens. By leveraging prompts, models can more easily focus on the interest targets, and prompt learning enables the model to learn complex visual concepts and capture discriminative spatio-temporal patterns effectively. Specifically, our prompts can be either predefined templates (non-learnable prompt: optical flow, large vision models) or learnable tokens (learnable prompt) that include task-specific information. They can be used either alone or in combination.

\subsubsection{Learnable Prompt: Soft Conditional Prompt Learning (SCP)}
\begin{figure}
    \centering
    \includegraphics[width=\columnwidth]{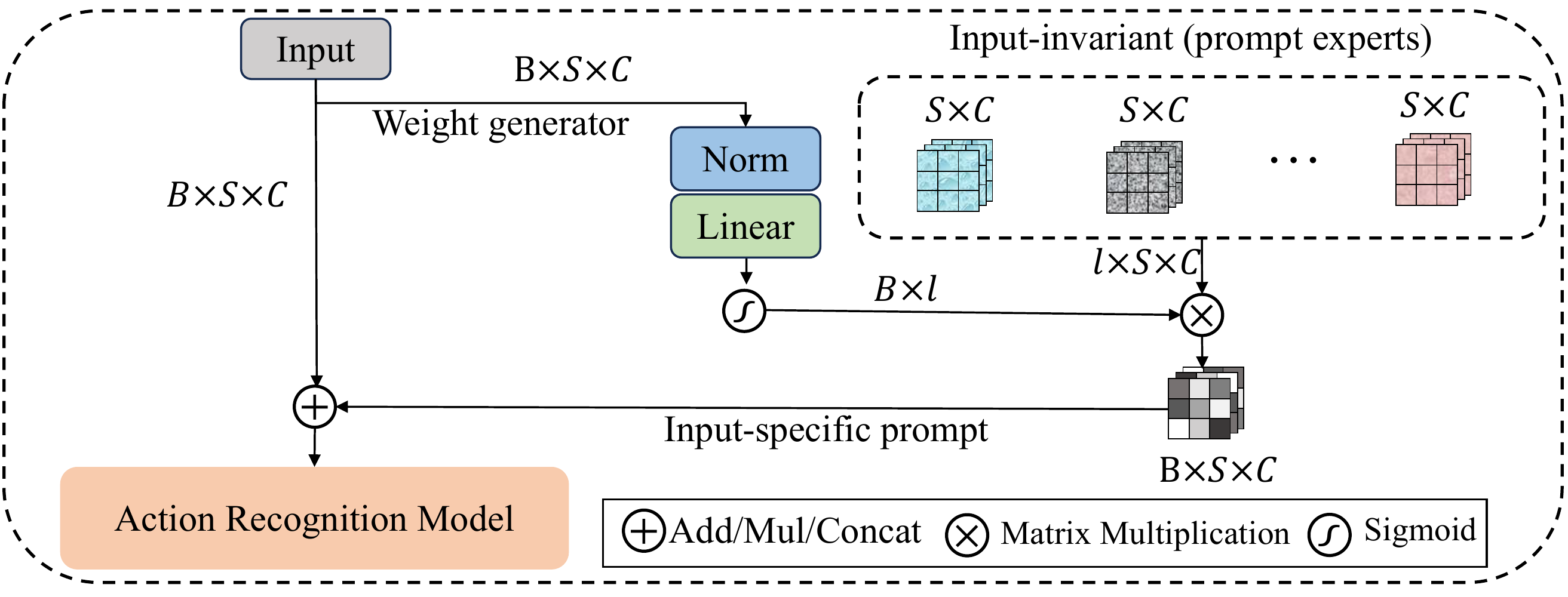}
    \caption{{\bf Soft Conditional Prompt Learning (SCP):} Learning input-invariant (prompt experts) and input-specific (data dependent) prompt. The input-invariant prompts will be updated from all the inputs, which contain task information, and we use a dynamic mechanism to generate input-specific prompts for different inputs. Add/Mul means element-wise operations. $B\times S\times C$ is the input features' shape, and $l$ is the expert's number in the prompt pool.}
    \label{fig:lp} 
    \vspace{-6mm}
\end{figure}

To better adapt to the input data, we propose a soft conditional prompt learning (SCP), which learns to dynamically generate prompts from a pool of prompt experts under different inputs. Prompt experts are learnable parameters that can be updated from the training process. As shown in Figure~\ref{fig:lp}, in our design, we use input-invariant (prompt experts) and input-specific (data dependent) prompts. The input-invariant prompts contain task information, and we use a dynamic mechanism to generate input-specific prompts for different inputs. 

There are different actions and domains (different video sources) for different videos, so it's challenging to learn a single general prompt for all videos. Therefore, we design an input-invariant prompt experts pool, which contains $l$ learnable prompts. Unless otherwise specified, the default value of $l$ is 8.
 \begin{equation}
P = \{P_1, ..., P_l\}, 
 \end{equation}
those prompt experts are learnable and will be updated from all the inputs. For a specific input $X^*$,
 \begin{equation}
P^* = Matmul(\sigma(FC(X^*)), P), 
 \end{equation}
We use an FC layer and sigmoid function $\sigma$ to get dynamic weights. Then we apply these dynamic weights to the input-invariant prompt pool to get a customized prompt $P^*$ for $X^*$.
 \begin{equation}
x_i^p= f_e([x_i,p_i]), x_i\in X^*,p_i \in P^*, 
 \end{equation}
where $x_i^p$ is the prompt-based feature.

\subsubsection{Non-Learnable Prompt}
Non-Learnable prompts make use of statistical methods (e.g., optical flow) or existing powerful large vision models, which can offer reliable prompts without training. 
\paragraph{Optical Flow Prompt}
Optical flow is a fundamental concept in computer vision that involves estimating the motion of objects within a video sequence. It represents the apparent motion of pixels between consecutive frames, providing valuable information about the movement of objects and their relative velocities. 

We divide a video into $m$ clips. For raw frame $x_i$ and frame $x_j$ from the video, the optical flow is:
 \begin{equation}
o_{i} = O(x_i, x_j), x_i\in clip_i, x_j\in clip_j,
 \end{equation}
where $clip_i$ and $clip_j$ are two adjacent clips from a video, and each clip contains several frames. When computing the optical flow, we only use one frame from each clip in a video and then apply the optical flow to this whole clip. This formulation is more efficient because it avoids many calculations for every frame. Therefore, the input with optical flow prompt becomes:
 \begin{equation}
[X,P] = \{x_k*o_{i} | \: x_k \in clip_i, i\in[1,m]\}
 \end{equation}
 where $clip_i$ has $k$ frames. We use $[X,P]$ to replace the original $X$ in video action recognition. 
 
\paragraph{Large Vision Model Prompt}

Recently, large models have been attracting more attention for NLP and other applications.  These large models are considered powerful since they are trained on huge amounts of data and don't need to be finetuned on new tasks as an auxiliary input (i.e. prompt). Our goal is to use these large models to generate prompts (e.g. mask, bbox) for video action recognition. 

One popular work is the Segment Anything Model (SAM~\cite{kirillov2023segment}), which can segment any object in an image given only some prompts like a single click or box. SAM is trained on a dataset of 11 million images and 1.1 billion masks. SAM can segment objects with high accuracy, even when they are new or have been modified from the training data. SAM generalizes to new objects and images without the need for additional training, so we don't need to finetune the model on our dataset. For some frames in a video clip, we generate a segmentation mask using a large vision model, SAM~\cite{kirillov2023segment}. Next, these masks are used as prompts and fused with input frames to optimize the recognition model. Specifically, for frame $x_i$, the output from SAM is:
 \begin{equation}
p_{i} = SAM(x_i, boxes/points), x_i\in clip_i
 \end{equation}
$clip_i$ is a video clip containing a few frames, 
 \begin{equation}
[X,P] = \{x_i*p_{i} | \: i\in[1,m]\}
 \end{equation}
 We use $[X,P]$ to replace the original $X$.

\subsection{Auto-regressive Temporal Reasoning}
Temporal reasoning is important for sequence data. Therefore, we propose an Auto-regressive Temporal Reasoning algorithm to better model the time-varying data. Auto-regressive models are statistical models that make predictions based on previous observations. They assume that the future values of a variable can be estimated by considering its past values. For temporal reasoning, this concept is extended to capture dependencies between different frames in a video.

After getting the prompt-based feature $X^p=\{x_1^p, x_2^p, ..., x_m^p\}$, where $x_i^p$ represents the observation at time step $i$, the goal is to predict the future values,
\begin{equation}
\hat{x}_{i+1}^p = f_a(\prod_j^{j<(i+1)}f_a(x_j^p)+x_{i+1}^p)
\end{equation}
where $f_a$ denotes the auto-regressive model that maintains an internal state and updates according to the sequential input. $\prod$ means a series of functions here. The auto-regressive temporal reasoning model considers the past observations of the sequence and the corresponding future observations to learn the underlying temporal dependencies. 

\subsection{Single-agent and Multi-agent Objective}
 The supervision formats used for single-agent and multi-agent action recognitions are different. As a result, we choose different loss functions. Specifically, we choose the classical cross-entropy loss for single-agent action recognition, 
\begin{equation}
  L_n=-\sum_{c=1}^C \log \frac{\exp \left(\hat{x}_{n, c}^p\right)}{\sum_{i=1}^C \exp \left(\hat{x}_{n, i}^p\right)} y_{n, c},  
\end{equation}
where $C$ is the class number, $n$ is the video number,  and $\hat{x}_{n, c}^p$ is the SCP's output feature. $y$ is the label. For multi-agent on Okutama, we use the BCEWithLogitsLoss,
\begin{equation}
    L_{n, c}=-\left[ y_{n, c} \cdot \log \sigma\left(\hat{x}_{n, c}^p\right)+\left(1-y_{n, c}\right) \cdot \log \left(1-\sigma\left(\hat{x}_{n, c}^p\right)\right)\right]
\end{equation}
where $\hat{x}_{n, c}^p$ is the SCP's output feature. $\sigma$ is a sigmoid function. This loss combines a sigmoid function and the BCELoss, which is more numerically stable than using a plain sigmoid followed by a BCELoss because by combining the operations into one layer, it takes advantage of the log-sum-exp for numerical stability. For both single-agent and multi-agent videos, by sharing the same objective, our learning approach can optimize prompts that guide the model's predictions while explicitly learning input-invariant (prompt experts pool) and input-specific (data-dependent) prompt knowledge. 

\section{Datasets and Results}
To verify the effectiveness of SCP, empirical evaluations were conducted on Okutama~\cite{barekatain2017okutama} and NEC Drone~\cite{choi2020unsupervised} comprising both single-agent and multi-agent actions. We further evaluate on Something-something V2~\cite{goyal2017something} ground camera videos to verify the effectiveness and generalization.


\begin{table}[t]
\centering
\begin{tabular}{c c c }
\toprule
Method & Frame size  & Accuracy    \\
\midrule
AARN \cite{yang2019framework, algamdi2020dronecaps} & crops  & $33.75\%$\\
Lite ECO \cite{zolfaghari2018eco, algamdi2020dronecaps} & crops  & $36.25\%$\\
I3D(RGB)\cite{carreira2017quo, algamdi2020dronecaps} &  crops & $38.12\%$\\
3DCapsNet-DR\cite{zhang2020capsnets, algamdi2020dronecaps} & crops  & $39.37\%$\\
3DCapsNet-EM\cite{zhang2020capsnets, algamdi2020dronecaps} & crops  & $41.87\%$\\
DroneCaps\cite{algamdi2020dronecaps} & crops  & $47.50\%$\\
DroneAttention without bbox\cite{yadav2023droneattention} & 720$\times$420 & $61.34\%$  \\
SCP without bbox (Ours) & 224$\times$224 & $71.54\%$  \\
\midrule
DroneAttention with bbox \cite{yadav2023droneattention} & 720$\times$420& $72.76\%$  \\
SCP with bbox (Ours) & 224$\times$224 & $75.93\%$  \\
\bottomrule
\end{tabular}
\caption{Comparison with the state-of-the-art results on the Okutama dataset. With bbox information, we achieved 10.20\% improvement over the SOTA method. Without bbox information, we outperformed the SOTA by 3.17\%. crops: from detection.}
\label{tab:okutama}
\vspace{-6mm}
\end{table}
\subsection{Datasets and Experiment Settings}
\paragraph{Okutama~\cite{barekatain2017okutama}} The Okutama dataset consists of 43 minute-long sequences with 12 action classes, providing a challenge with dynamic action transitions, changing scales and aspect ratios, camera movement, and multi-labeled actors. All the frames extracted from the video datasets were scaled to 224 × 224. The backbone is Swin-T~\cite{liu2021swin}. Following \cite{yadav2023droneattention}, the feature maps obtained were processed in the ROIAlign function (crop size of 5 × 5) to get the desired ROIs. Other training settings follow \cite{liu2021swin}.

\paragraph{NEC Drone~\cite{choi2020unsupervised}} features 5,250 videos depicting 16 distinct actions performed by 19 actors. The initial learning rate is set 0.05. Stochastic Gradient Descent (SGD) is used as the optimizer with 0.0005 weight decay and 0.9 momentum. We use cosine/poly annealing for learning rate decay. All the frames from the video datasets were scaled to 224 × 224.

\paragraph{Something-something v2 (SSV2~\cite{goyal2017something})} 
The SSV2 dataset is regarded as a substantial and comprehensive benchmark for action recognition, encompassing a vast collection of 220k action clips. Following ~\cite{li2022mvitv2}, we train for 100 epochs using 8 GPUs with a batch size of 64 and a base learning rate of 5e-5 with a cosine learning rate schedule. We use Adamw and use a weight decay of 1e-4 and a drop path rate of 0.4. For other training and testing settings, we follow ~\cite{li2022mvitv2}. And the backbone is MViTv2-S~\cite{li2022mvitv2}.





\subsection{Results on Okutama}
Okutama is an aerial multi-agent action recognition dataset in which multiple actors sequentially perform a diverse set of actions, which makes it very challenging. In the real world, it's difficult to ensure that only a single agent is in the scene for action recognition. Therefore, multi-agent action recognition is a very practical and important direction. We compare our SCP with state-of-the-art (SOTA) works.

As shown in Table~\ref{tab:okutama}, if there is no bbox information, we achieved 10.20\% improvement over the SOTA method. If there is bbox information, we outperform the SOTA by 3.17\%. This demonstrates the effectiveness of our method.

\subsection{Results on NECDrone}
We compare our method with other existing methods on NEC-Drone. The frames are extracted from raw videos and augmented as in X3D~\cite{feichtenhofer2020x3d}. The baseline methods use uniform and random sampling. As shown in Table~\ref{tab:nec}, on NEC Drone, our SCP outperforms the X3D by 4.0 - 7.4\% and improves 23.1\% over the K-centered.
\begin{table}[h]
\centering
\begin{tabular}{c c c}
\toprule
Method & pretrain  & Top-1 Accuracy   \\
\midrule
X3D-random \cite{feichtenhofer2020x3d} &  Kinetics400 & $52.0\%$  \\
X3D-uniform \cite{feichtenhofer2020x3d} &  Kinetics400 & $55.4\%$  \\
K-centered \cite{park2022k} &  N/A & $36.3\%$  \\ 
SCP-uniform (Ours) & Kinetics400 & $59.4\%$  \\
\bottomrule
\end{tabular}
\caption{Comparison with existing methods on NEC Drone. Our SCP improves 4.0-7.4\% over X3D and 23.1\% over K-centered.}
\label{tab:nec}
\vspace{-5mm}
\end{table}

\begin{figure*}
    \centering
    \includegraphics[width=\textwidth]{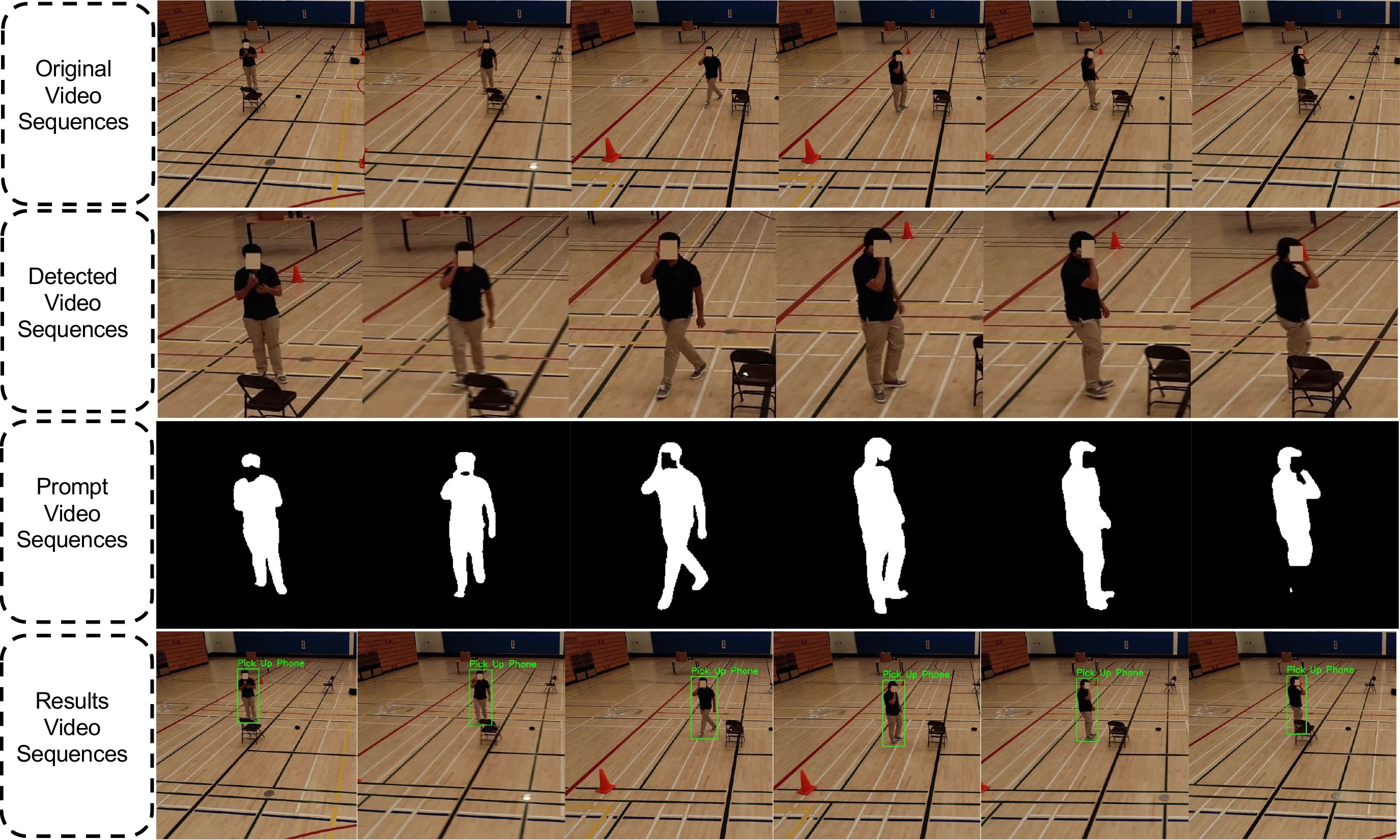}
    \caption{{\bf Visualization} We first detect the interested target and generate the prompts, then predict the action.}
    \label{fig:vis} 
\end{figure*}

\begin{table}[t]
\centering
\begin{tabular}{c c c c}
\toprule
Method & pretrain  & Top-1 Acc. & Top-5 Acc.    \\
\midrule
TEA \cite{li2020tea} &  ImageNet 1k & $65.1\%$ & $89.9\%$ \\
MoViNet-A3 \cite{kondratyuk2021movinets} &  N/A & $64.1\%$ &  $88.8\%$ \\
ViT-B-TimeSformer \cite{bertasius2021space} &  ImageNet 21k & $62.5\%$ & / \\ 
SlowFast R101, 8$\times$8 \cite{feichtenhofer2019slowfast} &  Kinetics400 & $63.1\%$ & $87.6\%$ \\ 
MViTv1-B, 16$\times$4 \cite{fan2021multiscale} &  Kinetics400 & $64.7\%$ & $89.2\%$ \\ 
MViTv2-S, 16$\times$4 \cite{li2022mvitv2} &  Kinetics400 & $67.3\%$ & $91.0\%$ \\ 
SCP (Ours) & Kinetics400 & $68.3\%$ & $91.4\%$ \\
\bottomrule
\end{tabular}
\caption{Comparison with the state-of-the-art results on the Something Something V2. Our SCP improves 3.6\% over MViTv1 and 1.0\% over strong SOTA MViTv2.}
\label{tab:ssv2}
\vspace{-1mm}
\end{table}
\subsection{Results on Something-something V2}
Something-something V2 is a challenging ground camera dataset for visual common sense because it requires models to understand the relationships between objects and actions. For example, to predict the category of a video, a model must understand that "something bounces a ball" is different from "something rolls a ball". 
We evaluate our SCP's reasoning and temporal modeling ability on Something-somethingV2.

As shown in Table~\ref{tab:ssv2},  our SCP improves 3.6\% over MViTv1 and 1.0\% over MViTv2, which illustrates the effectiveness of our proposed prompt learning and Auto-regressive temporal modeling.

\begin{table}[h]
\centering
\begin{tabular}{c c c }
\toprule
Component & Frame size  & Accuracy    \\
\midrule
Baseline & 224x224 & $71.54\%$  \\
Baseline + ROI  & 224x224 & $73.61\%$  \\
Baseline + ROI + Large Vision Model (SAM) & 224x224 & $74.68\%$  \\
Baseline + ROI + SCP & 224x224 & $76.34\%$  \\
\bottomrule
\end{tabular}
\caption{Ablation study in terms of the effect of different components in our method on the Okutama dataset. We evaluated ROI, Large Vision Model (SAM), and SCP. The experiments showed the effectiveness of our proposed methods.}
\label{tab:component}
\vspace{-1mm}
\end{table}
\begin{table}[h!]
\centering
\begin{tabular}{c c c }
\toprule
Method & Frame size  & Accuracy    \\
\midrule
Baseline & 224$\times$224 & $71.54\%$  \\
Baseline + Optical Flow & 224$\times$224 & $72.13\%$  \\
Baseline + Large Vision Model (SAM) & 224$\times$224 & $74.68\%$  \\
Baseline + SCP & 224$\times$224 & $75.93\%$  \\
\bottomrule
\end{tabular}
\caption{Ablation study in terms of different prompts on the Okutama dataset. We evaluated various prompts, including optical flow, a large vision model(SAM~\cite{kirillov2023segment}), and SCP. From our experiment, the large vision model and SCP achieved better accuracy.}
\label{tab:prompt}
\vspace{-2mm}
\end{table}

\subsection{Ablation Study}
First, we conducted ablation studies on various prompts, including optical flow, large vision models, and learnable prompts (SCP), to verify their effectiveness. Then we further evaluate the effect of each component of our method.


\textbf{Different Prompts}
To evaluate the effectiveness of different prompts, various prompts, including optical flow, large vision model (SAM~\cite{kirillov2023segment}), and learnable prompts, are examined in this work. As shown in Table~\ref{tab:prompt}, the large vision model and SCP achieved better accuracy.

\textbf{Effect of Each Component of Our Method}
We also evaluated the effect of the components in our methods, including Region of Interest alignment (ROI), Large Vision Model, and Learnable Prompt. As shown in Table~\ref{tab:component}, ROI can achieve 2.07\% improvement, ROI combined with Large Vision Model can achieve 3.14\% improvement, ROI combined with our SCP can achieve 4.80\% improvement. The experiments showed the effectiveness of our proposed methods.



\section{Conclusion}

We present a general prompt learning approach to alleviate the optimization burden by providing high-level texture descriptions or instructions associated with actions. 
These prompts enable the model to capture discriminative spatio-temporal patterns effectively. 
Our proposed SCP learns to dynamically generate prompts from a pool of prompt experts under different inputs. Our objective is to optimize prompts that guide the model's predictions while explicitly learning input-invariant (prompt experts) and input-specific (data-dependent) prompt knowledge. We observe good accuracy improvements on the challenging datasets. 


\textbf{Acknowledgement} This work was supported in part by ARO Grants  W911NF2310046 W911NF2310352  and U.S. Army Cooperative Agreement W911NF2120076.

\bibliographystyle{IEEEtran}
\bibliography{refs}

\end{document}